%% file: Main.tex
\documentclass[sigconf]{acmart}
\usepackage{algorithmic}
\usepackage{algorithm}
\usepackage{adjustbox}
\usepackage{graphicx}
\usepackage{orcidlink}
\usepackage{textcomp}
\usepackage{xcolor}
\usepackage{subcaption}
\usepackage{svg}
\usepackage{xkeyval}

\AtBeginDocument{%
  }

\setcopyright{acmlicensed}
\copyrightyear{2025}
\acmYear{2025}
\acmDOI{XXXXXXX.XXXXXXX}
\acmConference[EASE 2025]{The 29th International Conference on Evaluation and Assessment in Software Engineering}{17–20 June, 2025}{Istanbul, Türkiye}
\acmISBN{}

\begin{document}

\title{Testing Individual Fairness in Graph Neural Networks}

\author{Roya Nasiri}
\email{r.nasiri@tilbuurguniversity.edu}
\orcid{0009-0003-3985-3043}
\affiliation{%
  \institution{Jheronimus Academy of Data Science, Tilburg University}
  \city{’s-Hertogenbosch}
  \country{Netherlands}
}






\renewcommand{\shortauthors}{Nasiri et al.}

\begin{abstract}
The biases in artificial intelligence (AI) models can lead to automated decision-making processes that discriminate against groups and/or individuals based on sensitive properties such as gender and race. While there are many studies on diagnosing and mitigating biases in various AI models, there is little research on individual fairness in Graph Neural Networks (GNNs). Unlike traditional models, which treat data features independently and overlook their interrelationships, GNNs are designed to capture graph-based structures where nodes are interconnected. This relational approach enables GNNs to model complex dependencies, but it also means that biases can propagate through these connections, complicating the detection and mitigation of individual fairness violations. This PhD project aims to develop a testing framework to assess and ensure individual fairness in GNNs. It first systematically reviews the literature on individual fairness, categorizing existing approaches to define, measure, test, and mitigate model biases, creating a taxonomy of individual fairness. Next, the project will develop a framework for testing and ensuring fairness in GNNs by adapting and extending current fairness testing and mitigation techniques. The framework will be evaluated through industrial case studies, focusing on graph-based large language models.
\end{abstract}






\maketitle
  \textit{Advisors}: Willem-Jan van den Heuvel, Professor at Tilburg University (w.j.a.m.v.d.heuvel@uvt.nl) and Damian Andrew Tamburri,  Associate Professor at University of Sannio (datamburri@unisannio.it).
\input{Introduction}

\input{plan}

\input{progress}
\input{Related_Work}
\input{Conclusion}


\end{document}

%% file: Introduction.tex
\section{Introduction And Motivation}
\subsection{Context: Graph-structured Data based Applications}

Graph-structured data has applications across various domains, such as financial markets~\cite{Dayu2024finance}, social networks~\cite{Harsha2024SN}, recommendation systems~\cite{Jiawen2023recommneders}, and enterprise data analysis~\cite{Henna2021EnterpriseAU}. For example, graphs capture relationships between companies, investors, and assets in financial markets, aiding market trend analysis and risk management. In recommendation systems, graphs connect users, products, and preferences, enabling personalized suggestions. 

An emerging use case of graph-structured data is LLM (Large Language Model)-based applications~\cite{Perozzi2024LetYG}. Recent research indicates that integrating graph-structured data like knowledge graphs can significantly improve the performance of LLMs by enhancing their ability to understand and model complex relationships between entities~\cite{Perozzi2024LetYG, schneider2022decadeknowledgegraphsnatural, Pan2023UnifyingLL}. For instance, a Q\&A chatbot based on an enterprise knowledge graph can answer queries about business processes, customer data, or products, enhancing operational efficiency.

However, a key challenge for LLMs is the issue of hallucination~\cite{Wang2023SurveyOF}, where the models generate incorrect or nonsensical information that appears plausible. By incorporating graph-structured data, LLMs can more effectively access accurate relational information, improving their reasoning abilities, contextual understanding, and semantic knowledge extraction~\cite{fatemi2024talk}.

Graph-structured data is processed using various algorithms, including random walk-based methods~\cite{Barinov2015RW}, matrix factorization~\cite{Ziheng2018MF}, and Graph Neural Networks (GNNs)~\cite{Murgod2024surveygnn}. Among these, GNNs~\cite{Chengsheng2022GNN} have emerged as a key approach due to their ability to model complex relational structures and perform across diverse applications~\cite{Murgod2024surveygnn,Chengsheng2022GNN,Chen2024GNNsurvey,Zichong2023Gnnbias}.

GNNs capture relationships between entities by aggregating information from neighboring nodes. At each layer, a node updates its feature representation by combining its own features with those of its neighbors using an aggregation function and a learnable transformation matrix. This mechanism allows GNNs to effectively model interconnected data while preserving spatial and relational information~\cite{Dong2021REDRESS}. For example, in a social network, a user's profile is updated by aggregating features like interests, posts, or activity from their friends' profiles. This allows the model to learn from the user's connections, making better recommendations~\cite{Zhang2019GraphNN}.

Furthermore, GNNs can transform graph-structured data into sequences of graph tokens that carry essential structural information~\cite{Ren2024LLMGraphServey, Chengsheng2022GNN}. These tokens can then be integrated into LLMs, aligning the graph-structured data with natural language and improving performance, especially in tasks that involve complex relationships~\cite{Ren2024LLMGraphServey}.

However, as with other ML algorithms, it is critical to use GNNs ethically~\cite{Chen2024GNNsurvey}. For example, ML models are often perceived as more accurate, objective, and fair than human decision-making due to their ability to process vast amounts of data~\cite{mehrabi2022survey}. However, this belief is flawed, as machine learning algorithms can inherit and even amplify biases if not carefully designed~\cite{Pessach2023review,mehrabi2022survey}. For instance, Amazon’s ML hiring system discriminated against female candidates because it was trained on historically male-dominated data~\footnote{https://www.reuters.com/article/world/insight-amazon-scraps-secret-ai-recruiting-tool-that-showed-bias-against-women-idUSKCN1MK0AG/}. 

\subsection{Problem: Ensuring Fairness in GNNs}
GNNs are no exception, as biases present in graph structures and training data can propagate through model predictions, potentially leading to discrimination in critical decision-making~\cite{Zichong2023Gnnbias, Dai2020FairGNN}. 

Furthermore, the structure of graphs and the message-passing mechanisms of GNNs can amplify bias, as nodes with similar sensitive attributes tend to cluster, reinforcing disparities in decision-making. This results in highly correlated predictions with sensitive attributes, making GNNs more biased than models based solely on node features~\cite{Enyan2021sayno}. Furthermore, GNNs are designed to prioritize predictive accuracy, often at the expense of fairness. This focus on performance can inadvertently incorporate and perpetuate biases present in the data, undermining the fairness of GNNs in critical decision-making contexts.~\cite{Wang2024IF, Dong2023GM}. Such biases can hinder the fair application of GNNs in automated decision-making processes such as job matching~\cite{liu2021learning} and credit scoring~\cite{SHI2024gnncreditscoring}, where outcomes may have adverse personal and societal implications. 

Moreover, unlike independent and identically distributed (IID) data, where each instance is treated as independent, graph-based data is inherently non-IID, as nodes are interconnected and impact each other’s predictions through message-passing mechanisms in GNNs~\cite{Zhangnon-IIDgraph2025}. Consequently, fairness testing methods developed for IID settings~\cite{Zhang2020ADF, Zhang2021EIDIG, Wang2024MAFT} are ill-suited for graph-structured data, highlighting the need for GNN-specific fairness testing approaches.

Fairness in machine learning is typically defined through two primary notions: group fairness and individual fairness~\cite{Pessach2023review}. Group fairness aims to ensure equal treatment in different demographic groups~\cite{Hardt2016EqualityOO}, while individual fairness emphasizes that similar individuals should receive similar results~\cite{dwork2011fairnessawareness}. Although group fairness has been extensively studied, it does not always prevent discrimination within subgroups of protected classes or at the individual level~\cite{Debarghya2020twoIF, dwork2011fairnessawareness}. Furthermore, research indicates that different definitions of group fairness can sometimes be contradictory~\cite{Arrieta2019ExplainableAI, Kleinberg2016Inherentgf}. Consequently, recent ML fairness research has increasingly focused on individual fairness testing and mitigation~\cite{Li2023FL, Zhang2023ME, Wicker2023CertificationIF, Xiao2023LatentImitator, Wang2024IF}. 

While testing and ensuring individual fairness in ML models has made significant progress, research on individual fairness in GNNs remains in its early stages, with only a limited number of studies~\cite{Dong2021REDRESS, Fan2021FairGAE, Kang2020InFoRM, Liu2022IFMR}. Notably, no dedicated research on individual or group fairness testing specifically for GNNs~\cite{Chen2024FTsurvey}.

\subsection{Research Questions}
This research aims to address the existing gap in the literature regarding individual fairness in GNNs, as identified through my conducted systematic literature review (SLR) on individual fairness in AI systems. I introduce a framework to test and ensure individual fairness in these networks. This focus is essential because individual fairness, the principle that similar individuals should be treated similarly, is fundamental for achieving fairness, especially in areas where decision-making directly impacts individuals. One of the key challenges in fairness testing for ML models is generating test inputs that are both legitimate and natural~\cite{Chen2024FTsurvey}, adhering to real-world constraints and data distributions. To address this challenge, the framework proposed in this study provides a methodology for fairness testing in GNNs and emphasizes the generation of natural individual discriminatory instances. The framework can use the generated test cases to effectively identify fairness issues in GNNs and help mitigate them, all while maintaining model performance.

This Ph.D. project explores the following main research question.

\textbf{RQ: } \textit{How can individual fairness testing and mitigation practices be systematically applied to ensure fairness in GNNs?}

\noindent To explore this overarching question, I define two sub-questions that collectively shape and guide my investigation.

\textbf{SQ1: } \textit{What are the current approaches for testing and ensuring individual fairness in AI models?}

\noindent This involves systematically analyzing the literature on individual fairness in AI models to examine how existing practices and techniques define, measure, test, and mitigate individual fairness violations in them, leading to the development of an individual fairness taxonomy. It will also help identify challenges and gaps in ensuring individual fairness in GNNs. It will serve as a foundation for designing and implementing a framework for individual fairness testing and ensuring in GNNs. 

\textbf{SQ2: } \textit{How can existing individual fairness techniques be adapted and extended to test and improve individual fairness in GNNs effectively and efficiently?}

\noindent This question will focus on adapting and extending existing fairness testing and mitigation methods to address fairness challenges in GNNs. The goal is to effectively generate individual discriminatory instances, which involves creating natural and realistic test cases that accurately capture fairness issues. At the same time, the techniques will be applied efficiently, meaning that fairness testing and mitigation should not compromise the model's performance, ensuring an optimal balance between fairness and model effectiveness. The proposed individual fairness testing framework will be evaluated through experiments in both synthesized and real-world scenarios to demonstrate its effectiveness and efficiency.

\textbf{RQ2: }\textit{To what extent can the proposed individual fairness framework in GNNs ensure fairness in industrial applications?} 

This research is conducted in collaboration with Deloitte’s
Trustworthy AI team\footnote{https://www2.deloitte.com/us/en/pages/deloitte-analytics/solutions/ethics-of-ai-framework.html}. By following an action research approach
~\cite{Staron2020ActionResearch}, the proposed framework is iteratively tested and refined on industrial GNN applications, with a particular focus on graph-based LLMs, which Deloitte uses in its internal applications, to assess its effectiveness in detecting and mitigating individual fairness issues.






%% file: plan.tex
\section{Research Plan}
\begin{figure*}[!ht]
\centering
\includegraphics[scale=0.75]{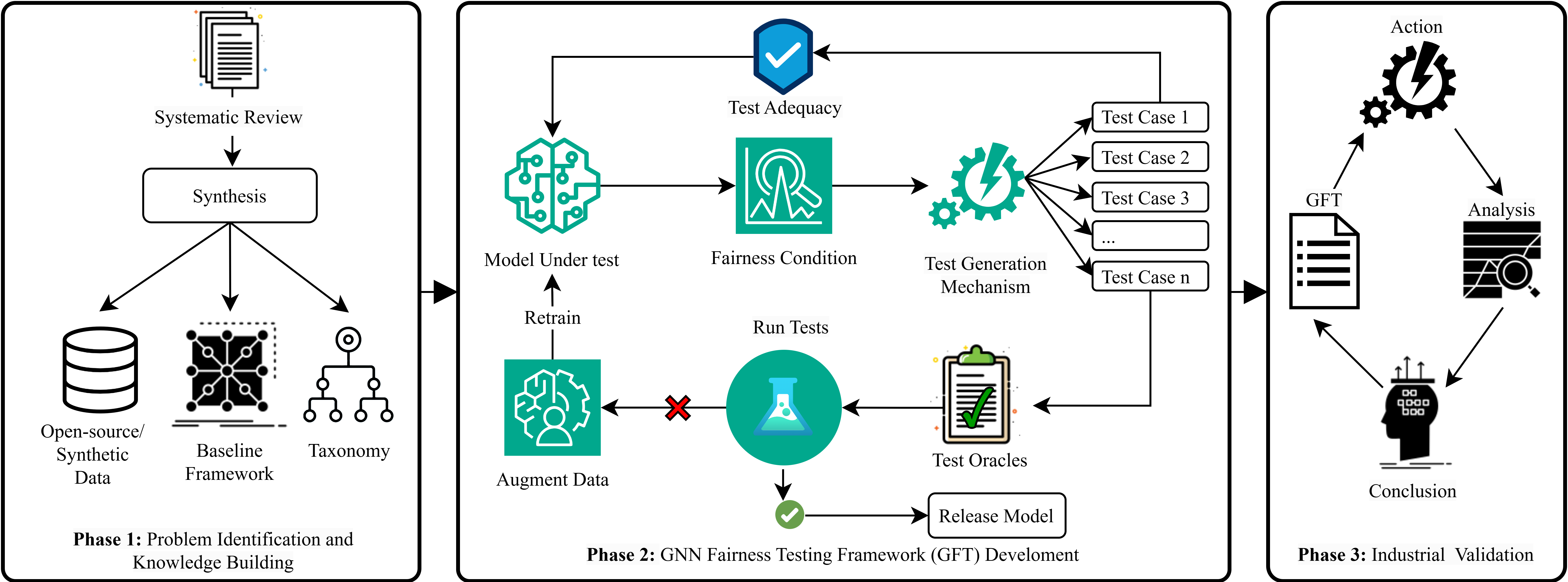}
\centering
\caption{A Framework for Testing and Ensuring Individual Fairness in GNNs}
\label{fig:framework}
\end{figure*}

The project adheres to the design science research methodology, which focuses on identifying problems, developing solution artifacts, and validating them within a real-world context~\cite{Peffers01122007,Runeson2020DSP}. The research is organized into three phases, as illustrated in Figure~\ref {fig:framework}.

\begin{figure*}[!ht]
\centering
\includegraphics[scale=0.20]{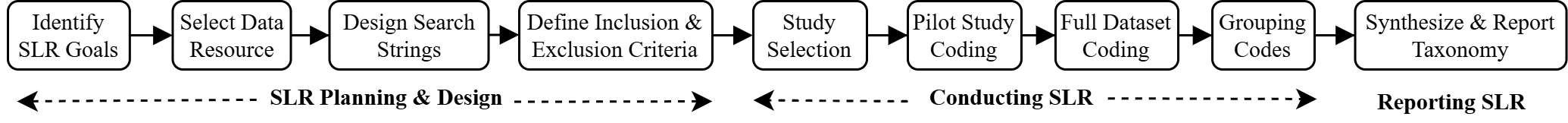}
\centering
\caption{Overview of SLR Process}
\label{fig:slr}
\end{figure*}

\textbf{Phase 1: Addressing \textbf{SQ1}}: A systematic literature review (SLR) was conducted to identify research challenges and gaps in auditing individual fairness in AI models. The methodology follows the SLR guidelines commonly used in software engineering~\cite{Garousi2017SLRGuidence} and aligns with those used in other SLR studies~\cite{Islam2019Multivocal}. The review process consists of three phases: planning, conducting, and reporting. During the planning phase, research objectives are defined, and a protocol is developed to guide the review's focus and methodology. The conducting phase includes identifying and selecting relevant studies on individual fairness in AI systems, followed by a thematic analysis of the collected data (Figure~\ref{fig:slr}).

This review aims to synthesize a taxonomy of definitions, metrics, testing approaches, and mitigation strategies for individual fairness. It will provide valuable insights into fairness testing and test generation techniques, offering a deeper understanding of the evolution of individual fairness, its primary applications, and the challenges encountered across various domains.

The SLR indicates that although fairness has been examined in various AI domains, fairness testing in GNNs remains underexplored, likely due to their intricate structures and the challenges in defining fairness metrics. This gap emphasizes the need for methods to evaluate and address individual fairness in GNNs.

\textbf{Phase 2: Addressing \textbf{SQ2}}: This phase develops a fairness testing framework for GNNs that consists of five key stages: test input generation, test adequacy evaluation, test oracle generation, test execution, and model retraining.
First, I will do a deep understanding of GNN architectures, focusing on how graph topology, neighborhood aggregation, and structural imbalances in graph-structured data contribute to unfairness in GNNs. These insights guide the adaptation of fairness testing techniques (e.g., gradient-guided adversarial sampling~\cite{Wang2024MAFT}, GANs~\cite{Xiao2023LatentImitator}) to generate natural individual discriminatory instances (IDIs) for GNNs. Unlike traditional models, generating IDIs for GNNs requires preserving both the node’s features and the structural dependencies induced by the graph topology. Therefore, existing techniques will be extended to operate over graph-structured data by incorporating topological constraints, such as preserving node degrees and neighborhood distributions. 
In Fairness testing, an individual discriminatory instance (IDI) is defined as an input where the model’s prediction changes solely due to variations in a protected attribute (e.g., gender, race, or age) while all other attributes remain unchanged~\cite{Wang2024MAFT}. By generating and testing IDIs, potential fairness violations in an ML model can be detected~\cite{Zhang2021EIDIG}. For example, consider a dataset with 10 attributes, where gender is the protected attribute (the 2nd attribute). Suppose we have two instances:
\[
x = [0, \textcolor{red}{1}, 30, 1, 2, 0, 0, 1, 1, 0]  \quad \textrm{and} \quad x' = [0, \textcolor{red}{0}, 30, 1, 2, 0, 0, 1, 1, 0].
\]
Here, $x$ and $x'$ are identical in all attributes except for gender. If these instances exist and the model predicts differently for them, $x$ is considered a discriminatory instance regarding gender.

The framework will generate these cases and ensure their naturalness and validity by adhering to the original data distribution. The test cases will then be verified using test oracles. A test oracle in fairness testing determines whether the model adheres to fairness requirements and helps identify fairness violations~\cite{Chen2024FTsurvey}. Statistical parity difference, for instance, detects fairness violations by comparing the favorable outcome rates between groups (e.g., males vs. females). Such violations guide mitigation, like retraining with generated IDIs to reduce bias. This ensures the model's fairness while maintaining overall performance~\cite{Zhang2021FairnessTO, Xiao2023LIMI, Zhang2020ADF}. Test adequacy will be assessed to ensure fairness evaluations are complete and representative~\cite{Zhang2021FairnessTO, Chen2024FTsurvey}. For this, I propose a GNN-specific adequacy metric that extends fairness neuron coverage, measuring how well test cases activate fairness-relevant components in graph-based models.

Test case generation techniques face limitations such as high complexity and scalability issues. While parallelization and targeted sample selection~\cite{Zhang2021FairnessTO, Tao2022Ruler} can help mitigate these issues, fully resolving them is beyond the scope of this project. The focus here is on adapting these techniques to graph-structured data, an area where this work remains limited.

\textbf{Phase 3: Addressing \textbf{RQ2}}: This phase will follow an action research methodology~\cite{Holstein019FairML, Staron2020ActionResearch, Pfeuffer2023XAIRS}, using iterative problem-solving and collaboration to bridge theory and practice. In partnership with Deloitte, the study will apply the proposed fairness testing framework to detect and mitigate individual fairness violations in real-world GNN applications in Deloitte. The framework will be evaluated using both real-world and synthetic datasets to assess its effectiveness in identifying model biases and their underlying causes. The evaluation will help refine fairness testing strategies and provide insights for integrating them into Deloitte's audit workflows.

\subsection{Data Collection}

Based on the literature review conducted in Phase 1, open-source and synthetic datasets, such as the German~\footnote{https://archive.ics.uci.edu/dataset/144/statlog+german+credit+data}and Credit~\cite{Yeh2009credit} datasets, are commonly used in fairness research and can be utilized in our study. The German dataset contains credit information from a German bank, where each client is represented as a node in a graph, and edges indicate the similarity between clients' credit accounts. Gender is considered a sensitive attribute to classify clients into good or bad credit risk categories~\cite{Wang2023MMB}. Moreover, the industrial partner will support us by providing anonymized data and knowledge graphs from their clients.

However, recent research indicates that poorly constructed semi-synthetic datasets, converted from tabular data and various real-world datasets, present a significant issue in the fairness domain of GNNs~\cite{Chen2024GNNsurvey, Qian2024Data}. The primary concern is that the graph structure of these constructed datasets fails to provide additional meaningful information in the edges, as they are built based on feature similarity rather than capturing more complex relational data~\cite{Qian2024Data}.

Fortunately, synthetic generation methods exist to address these issues, including the approach proposed by Qian et al.~\cite{Qian2024Data}. I will utilize theirs and other relevant approaches to ensure that the training data captures richer relational data and enhances the fairness of the GNN models I am developing. 





%% file: progress.tex
\section{Progress So Far}
I conducted an SLR to identify research challenges and gaps in auditing individual fairness in AI models, following Kitchenham’s guidelines~\cite{kitchenham2004procedures}. The search across five digital libraries (IEEE Xplore, ACM, Scopus, Web of Science, DBLP) initially yielded 1,118 papers, with 136 selected after applying inclusion criteria. 35\% were from journals and 65\% from conferences. The papers were thematically coded using Atlas.ti~\footnote{https://atlasti.com/}, and I am developing an individual fairness taxonomy to classify definitions, metrics, testing methods, mitigation strategies, and challenges.

The industrial partner of the PhD project and I also agreed on an industrial use case for this project. The company constructs knowledge graphs from reports and integrates them into a Graph-based LLM approach~\cite{Huang2024GNNLLM} to answer employee queries. However, the current performance shows bias issues and hallucinations. A potential enhancement involves leveraging GNNs to learn more robust knowledge graph representations, which can then be integrated with LLMs to improve response accuracy. This research will explore how the proposed fairness testing framework can be effectively applied to GNNs in this context, ensuring fair, reliable, and high-quality outcomes aligned with the company’s objectives.

%% file: Related_Work.tex
\section{Related Work}
This section first reviews the literature on fairness testing. Next, it discusses the fairness in GNN, highlighting the research gaps related to individual fairness testing. 

\subsection{Fairness Testing}

Fairness is a key requirement in ML development, making fairness testing essential~\cite{Zhang2022MLTesting}. Several recent surveys address fairness in ML models and techniques for testing fairness~\cite{Chen2024FTsurvey, mehrabi2022survey, kheya2024pursuitfairness}. Based on these surveys and the SLR, individual fairness testing approaches can be divided into black-box and white-box methods.

Black-box fairness testing operates with limited knowledge of the model. Key studies include THEMIS~\cite{Galhotra2017Themis}, which generates test cases using the random sampling strategy to detect discrimination. AEQUITAS~\cite{Udeshi2018Aequitas}, which refines this using a two-phase search strategy to identify discriminatory seeds; and SG~\cite{Aggarwal2019SG}, which combines symbolic execution with local explainability to detect discrimination. These methods are effective but computationally intensive and generate duplicates~\cite{Chen2024FTsurvey}.


White-box fairness testing, with full model access, includes ADF~\cite{Zhang2020ADF}, which uses gradient-based sampling near decision boundaries; NeuronFair~\cite{Zheng2022NeuronFair}, which targets biased neurons to mitigate gradient vanishing in ADF; and DiCE~\cite{Monjezi2023DiCE}, which applies gradient-guided clustering. Despite these advances, they often fail to generate natural discriminatory instances, limiting their effectiveness~\cite{Tao2022Ruler}.

Recent advancements in test generation include ExpGA~\cite{Fan2022ExpGA}, which combines interpretable models with a genetic algorithm, and LIMI~\cite{Xiao2023LIMI}, which uses GANs to enhance the naturalness and efficiency of generated instances~\cite{Chen2024FTsurvey}.

\subsection{Fairness in GNNs}
GNNs are widely applied in graph-based learning tasks, and fairness research has primarily addressed group fairness in GNNs through bias mitigation strategies. According to a survey on fairness in GNNs~\cite{Chen2024GNNsurvey}, these mitigation techniques can be categorized based on their approach into fairness-aware random walks, adversarial debiasing, Bayesian methods, and optimization-based regularization. For example, FairGNN ensures fair node classification via adversarial learning~\cite{Dai2020FairGNN}, 
while Monet prevents metadata leakage with orthogonal embeddings~\cite{Palowitch2021MONET}.

Research on individual fairness in GNNs is limited. REDRESS is a ranking-based framework that promotes fairness while maximizing GNN utility~\cite{Dong2021REDRESS}. IFMR is a multi-similarity fairness loss function specifically designed for medical recommendations~\cite{Liu2022IFMR}. FairGAE explores fairness-aware graph representations using Wasserstein distance to learn fair graph embeddings~\cite{Fan2021FairGAE}. InFoRM offers a general framework for individual fairness across graph mining tasks, supporting debiasing through graph modifications, model adjustments, and result refinement~\cite{Kang2020InFoRM}.


While most existing methods~\cite{Zhang2020ADF, Zhang2021EIDIG, Monjezi2023DiCE, Tao2022Ruler, Wang2024MAFT} assume IID data and treat each instance independently (or only perturb feature vectors in isolation), they cannot capture the non-IID, message-passing behavior of GNNs~\cite{Zhangnon-IIDgraph2025}. This PhD study adapts fairness testing frameworks to GNNs by proposing a tailored GNN-specific test case generation technique that applies structure-preserving perturbations, maintaining node degree and neighborhood consistency. It also introduces a novel test adequacy criterion based on layer-wise fairness neuron coverage. These contributions enable effective fairness testing in GNNs, beyond the capabilities of IID-based approaches.

%% file: Conclusion.tex
\section{Conclusion}
This paper discussed the research plan and status of my PhD. It presented the research objectives, methodologies, evaluation plans, expected challenges, and progress. The research plan for my PhD is comprised of three phases. I completed the first year of my PhD trajectory (March 2025) and made substantial progress in the first phase. In particular, I have completed a systematic literature review (SLR) that provides a taxonomy for individual fairness and identifies key challenges and gaps in fairness testing for GNNs. In the section phase, the main goal is to develop a framework for individual fairness testing in GNNs. The anticipated outcome is a comprehensive toolkit that adapts existing fairness testing techniques to GNN-specific characteristics, enabling the detection and mitigation of fairness violations in GNNs. This framework will offer AI practitioners practical guidelines to assess and improve fairness in GNN models, ensuring that model performance is maintained. Additionally, it will provide actionable insights for integrating fairness testing into AI system audits through real-world validation and industry collaboration, enhancing transparency and trust in GNN-based applications (the third phase).